\documentclass[10pt,twocolumn,letterpaper]{article}

\usepackage{wacv}
\usepackage{times}
\usepackage{epsfig}
\usepackage{graphicx}
\usepackage{amsmath}
\usepackage{amssymb}

\usepackage[numbers]{natbib}
\usepackage{stmaryrd}
\usepackage{makecell}
\usepackage{multirow}
\usepackage{sg-macros}
\usepackage{makecell}
\usepackage{spreadtab}
\usepackage{authblk}

\newcommand{\ops}{\mathcal{O}}

\usepackage[pagebackref=true,breaklinks=true,letterpaper=true,colorlinks,bookmarks=false]{hyperref}

\wacvfinalcopy 
\def\wacvPaperID{553} 

\ifwacvfinal\fi
\begin{document}
\title{Neural Algebra of Classifiers}

\author[1]{Rodrigo Santa Cruz}
\author[1]{Basura Fernando}
\author[1,2]{Anoop Cherian}
\author[1]{Stephen Gould}
\affil[1]{Australian Centre for Robotic Vision, Australian National University, Canberra, Australia}
\affil[2]{Mitsubishi Electric Research Labs, 201 Broadway, Cambridge, MA}
\affil[ ]{\tt\small{firstname.lastname@anu.edu.au}}

\maketitle
\ifwacvfinal\thispagestyle{empty}\fi

\begin{abstract}
The world is fundamentally compositional, so it is natural to think of visual recognition as the recognition of basic visually primitives that are composed according to well-defined rules. This strategy allows us to recognize unseen complex concepts from simple visual primitives. However, the current trend in visual recognition follows a data greedy approach where huge amounts of data are required to learn models for any desired visual concept. In this paper, we build on the compositionality principle and develop an ``algebra'' to compose classifiers for complex visual concepts. To this end, we learn neural network modules to perform boolean algebra operations on simple visual classifiers. Since these modules form a complete functional set, a classifier for any complex visual concept defined as a boolean expression of primitives can be obtained by recursively applying the learned modules, even if we do not have a single training sample. As our experiments show, using such a framework, we can compose classifiers for complex visual concepts outperforming standard baselines on two well-known visual recognition benchmarks. Finally, we present a qualitative analysis of our method and its properties.
\end{abstract}

\section{Introduction}
Imagine a sea-faring bird with ``hooked beak'' and ``large wingspan''. Most people would be thinking of an albatross. Moreover, given a set of images of birds, the descriptive features ``hooked beak'' and ``large wingspan'' are key for someone to identify images of albatross versus other birds even if they had never seen an albatross before. These provide evidence that visual concepts are compositional and complex visual concepts like albatross are defined as a composition of simpler visual concepts such as ``hooked beak'' and ``large wingspan''. In addition, humans have very formal and structured ways of reasoning about compositions such as propositional logic, predicate logic, and boolean algebra. However, the current state-of-the-art models for recognition follow a laborious data-driven approach, where complex concepts are learned using thousands or millions of manually labeled examples instead of using composition. Such data greedy approach is infeasible for many real world applications.

\begin{figure}[t]
	\begin{center}		
		\includegraphics[width=0.47\textwidth]{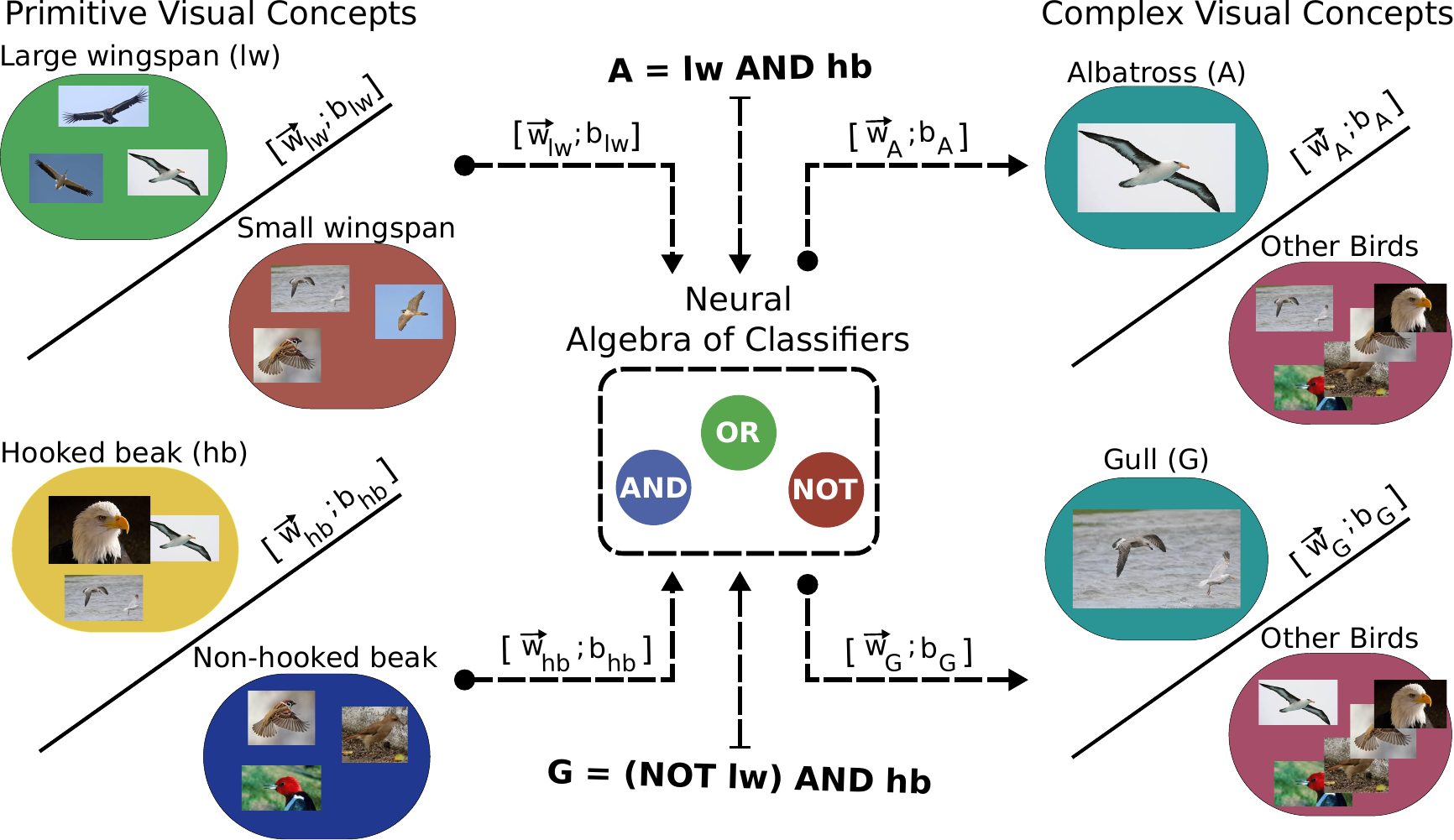}		
	\end{center}
	\label{fig:intro}
    \vspace{-5px}
	\caption{Illustration of the proposed neural algebra of classifiers. Given classifiers for primitive visual concepts such as hooked beak and large wingspan, we can compose classifiers for complex concepts such as gull and albatross that are represented by boolean expressions of these primitives.}
\vspace{-10px}
\end{figure} 

In this paper, we build on the insight that visual concepts are fundamentally compositional and develop an algebra for combining concept classifiers. Towards this end, we propose a composition framework inspired by boolean algebra structures such as disjunction, conjunction, and negation. More specifically, we develop neural network modules which can learn to compose classifiers according these logical operators allowing us to produce classifiers for any complex concept expressed as a boolean expression of primitive concepts. For instance, our approach can compose a classifier for albatross by combining classifiers for ``large wingspan AND hooked beak''. Likewise, gull's classifier can be expressed as ``(NOT large wingspan) AND hooked beak'' (\figref{fig:intro}). Moreover, such a framework can predict unseen complex visual concepts like humans do. For example, it is possible to identify a car made of grass by composing a classifier for ``grass AND car'', even if such a concept does not have training data. It also allows to recognize subclasses and specific instances of objects without any additional annotation effort. Therefore, we can scale-up recognition systems for complex and dynamic scenarios.

Learning how to compose classifiers for unseen complex concepts from simple visual primitives by developing a compositional algebra is a challenging task since there is no trivial mapping between primitives and their compositions. Naively, we can think of recognizing an albatross whenever the classifiers for large wingspan and hooked beak fire simultaneously. However, such an approach assumes strong independence between visual primitives and does not consider the imperfection of the primitive classifiers or reason about correlations and cooccurrences of visual primitives. Furthermore, as observed by \citet{Misra:Composing:CVPR2017}, the meaning of a composition depends on the context and the particular instance being composed. For instance, the visual appearance of ``old'' for bikes is completely different for people. In contrast, our approach is learned in the classifier space exploring correlations, cooccurrences, and contextuality between visual primitives in order to compose more accurate classifiers for complex visual concepts.

Our contributions are threefold. First, we propose a learning framework for composition of classifiers. Such a framework resembles an algebra in which we can synthesize classifiers for any visual concept described as boolean expression of visual primitives. 
Second, we develop a neural network based model which minimizes the classification error of a subset of visual compositions and generalizes for unseen compositions. Third, we show how these modules can be used recursively to produce classifiers for complex concepts expressed as boolean expressions of visual primitives.  

We conduct several experiments to show the efficacy of our approach. We show that our method is able to synthesize classifiers according to simple composition rules by learning how to compose concepts from a subset of primitive compositions and generalizing for compositions not seen during training. In addition, our approach naturally extends to complex compositions by recursively applying our learned neural network modules. On all of these settings, our method outperforms standard baselines. Finally, we evaluate qualitatively some interesting properties of our method.

\section{Related Work \label{sec:relw}}
The principle of compositionality says that the meaning of a complex concept is determined by the meanings of its constituent concepts and the rules used to combine them \citep{Frege:1948,Boole:1854,Burnyeat:1990}. For instance, written language is built of symbols which form syllables, words, sentences, and texts. Likewise, visual data can be decomposed into scene, objects, textures and pixels. The principle is pervasive in our world and have been studied extensively by different scientific communities ranging from mathematics to philosophy of language. In this paper, we study compositionality in the context of visual recognition.
 
Viewing objects as collections of known parts at familiar relative locations may be the most common way to incorporate compositionality into visual recognition systems. For instance, deformable parts model \citep{Felzenszwalb:PAMI2010,Girshick:NIPS2011}, and-or graphs \citep{Wu:IJCV2011,Si:PAMI2013,Zhu:CVPR2008,Tang:ICCV17}, dictionary learning \citep{Tu:IJCV2005,Zhu:CVPR2010,Zhu:2007}, and self-supervised representation learning \citep{Doersch:ICCV2015,SantaCruz:CVPR2017,Fernando:CVPR2017} techniques are built over this intuition. Likewise, scenes can be seen as hierarchical compositions of concepts in different abstraction levels. Then, convolutional neural networks \citep{Zeiler:ECCV2014, Simonyan:2014} and recurrent neural networks \citep{Hochreiter:1997,Chung:2014,Socher:ICML2011} can also be seen as compositional models. Differently, we focus in composing classifiers for complex concepts that can be expressed as boolean expression of primitive visual concepts. For instance, our approach is able to classify a specific instance given its visual attributes even if such an instance is not present in the training set.

It is important to note that compositionality helps to reduce the complexity of some problems by decomposing them in subproblems which allow more tractable solutions. For instance, \citet{Andreas:CVPR2016} and \citet{Hu:ICCV17} explore the structure of natural language questions in order to define a set of simpler problems which can be solved by simple neural networks. \citet{Neelakantan:ICLR16}, proposed a neural network to induce programs of simple operations to answer questions which involve logic and arithmetic reasoning. \citet{Faktor:ECCV12,Faktor:CVPR2013} use the ``similarity by composition'' framework \citep{Boiman:NIPS2007} to perform clustering and object co-segmentation. Likewise, we decompose the problem of recognizing any specific instances of objects by the problem of composing a classifier according to simple rules from its individual visual primitives.

Closely related to our work, \citet{Misra:Composing:CVPR2017} show the importance of context in composition of object and attributes. More specifically, they argue that the visual interpretation of attributes depends on the objects they are coupled with. For instance, an old bike has different visual features than an old computer. Building on this intuition, the authors propose a transformation function to map from object and attribute classifiers to the composition of classifiers. Thus, their scheme can only synthesize classifiers for visual concepts like ``red wine'', ``large tv'', and ``small modern cellphone''. In contrast, we develop a generic framework to combine any number of concept classifiers according to arbitrary boolean expressions. Such a framework provides richer expressiveness since we are able to compose classifiers for more complex concepts like ``red or blue socks without holes''.

The problem of classifying unseen visual concepts is also known as zero-shot classification \citep{Palatucci:NIPS2009,Lampert:CVPR2009,Lei:CVPR2015,Frome:NIPS2013}. However, zero-shot classifiers are only able to recognize unseen object classes, while our proposed framework is also able to recognize unseen groups, sub-groups, and specific instances of objects. Furthermore, we do not make assumptions about the existence of an external source of knowledge such as class-attributes relationship \citep{Lampert:CVPR2009}, text corpus \citep{Lei:CVPR2015}, or language models \citep{Frome:NIPS2013}. We explore compositionality in the visual domain and other visual priors, such as co-occurrence and dependence of visual attributes.

\section{Neural Algebra of Classifiers \label{sec:method}}
In this section, we explain the proposed neural algebra of classifiers. We start by formalizing the problem of classifier composition in an algebraic perspective. Then, we describe our learning algorithm, model architecture, and inference pipeline.
\begin{figure*}[t!h]
	\begin{center}		
		\includegraphics[width=\textwidth]{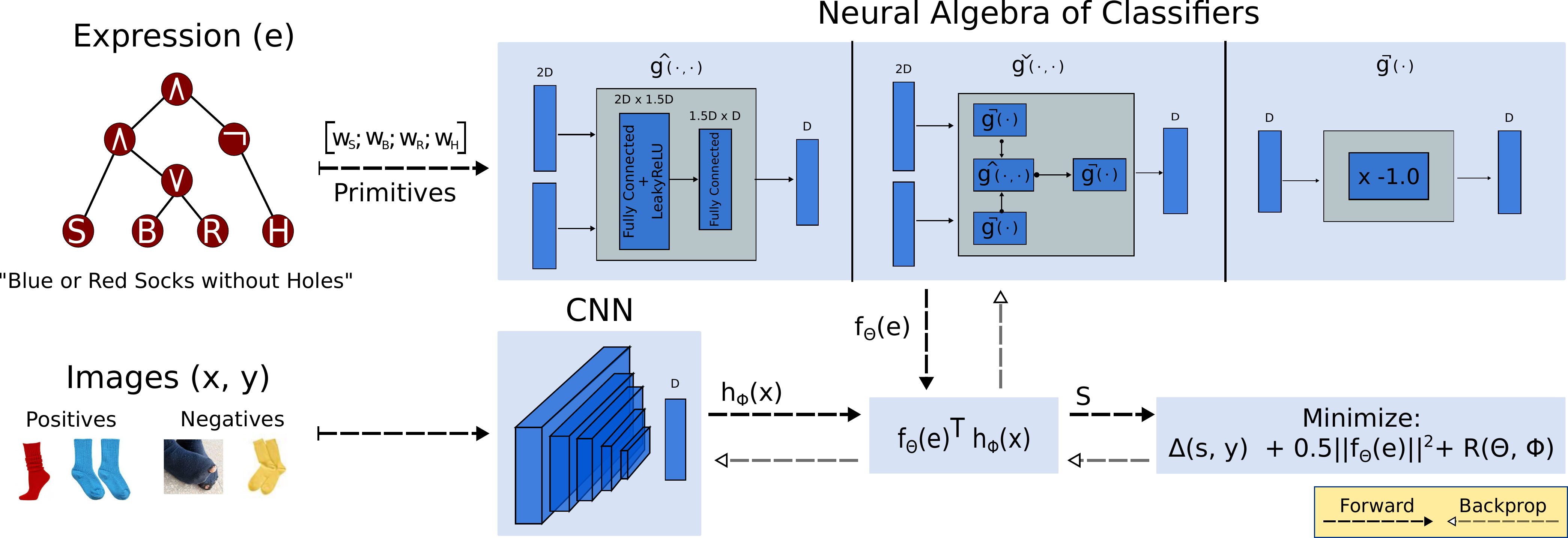}
	\end{center}
\vspace{-10px}
	\caption{Neural algebra of classifiers: Our method composes classifiers for complex visual concepts expressed as boolean expressions of primitive concepts. We first represent every primitive by a classifier and every image by a feature vector. Then, we use a subset of training expressions to learn a set of composition functions which generalizes to concepts represented by unseen expressions or even unseen primitives. In order to make such learning problem easier, we explore geometry and boolean algebra fundamentals such as hyperplanes and De Morgan's laws.}
\label{fig:model}
\vspace{-10px}
\end{figure*}

\subsection{Problem Formulation}
Our problem consists of classifying images according to complex visual concepts expressed as boolean algebra of a set of primitives. Initially, let us assume we have a set of known visual concepts, named primitives, like socks (S), Red (R), Blue (B) and Holes (H). In addition, consider basic composition rules inspired by boolean operators: ($\land$) that identifies whether two primitives are depicted in the image simultaneously, ($\lor$) which denotes if the image has at least one of the primitives, and ($\neg$) which accepts all images which a primitive is not depicted. Then, what is the classifier for a complex visual concept expressed by multiple compositions of primitives and these rules. For instance, what is the classifier for ``red or blue socks without holes'' described by the expression ``S $\land$ (B $\lor$ R) $\land$ ($\neg$ H)''.

Formally, let us define a set of \emph{primitives} $\P=\{p_i\}_{i=1}^M$. We can express complex concepts by forming arbitrary \emph{expressions} recursively combining primitives with \emph{composition rules} $\ops =\{\neg, \land, \lor\}$. Note that this set of rules is a complete functional set, i.e., any propositional expression of primitives can be written in terms of these rules. Then, our objective can be summarized as learning a parametrized function, $f_\theta\func: \E \mapsto \C$ that maps from the space of expressions $\E$ to a space of binary classifiers $\C$. In other words, we want the function $f_\theta\func$ be able to synthesize a classifier for any given expression. 

Without loss of generality, we will explain the details of our approach for the case of linear classifiers, but the same formulation can be used to synthesize non-linear or kernelized classifiers. Thus, we define $f_\theta\func$ as,
\begin{equation}
\hat{w}_{e} = f_\theta(e)
\end{equation}
where $\hat{w}_{e} \in \C$ is a linear classifier, i.e., separating hyperplane, that distinguishes positive and negative samples for an expression and $\theta$ are the function parameters.

\subsection{Learning}
In order to efficiently learn the proposed mapping function, we need to represent the visual content of images and the semantic meaning of primitives in a compact way. Towards this end, we define $h_{\phi} \in \reals^D$ as a parametrized feature extractor which computes a vector representation that summarizes all visual features of a given image and $\phi$ is the set of parameters. Likewise, we represent all primitives by classifiers trained to recognize images that depict them. Since we focus on linear classifiers in this paper, we represent every primitive $p$ by the separating hyperplane parameters $w_p \in \reals^{D}$, e.g., obtained by training an one-vs-all linear SVM classifier on the feature representation of images.

Note that boolean expressions are evaluated by decomposing them into a sequence of simpler terms and evaluating these terms recursively. For instance, the expression $S \land (B \lor R) \land (\neg H)$ can be evaluated by recursively evaluating the sequence of simpler expressions $(B \lor R)$, $S \land (B \lor R)$, $\neg H$, $((S \land (B \lor R)) \land (\neg H))$. Such a decomposition can be computed efficiently by representing expressions as binary trees and parsing their nodes in post-order. Then, we propose to model the function $f_\theta\func$ as a set of composition functions $g^{\land}\func$, $g^{\lor}\func$, $g^{\neg}\func$. In other words, the function $f_{\theta}\func$ is computed by decomposing an expression in simple terms and applying the composition functions accordingly. 

These composition functions $g^*:\C \times \C \mapsto \C$ are auto-regressive models which maps from and to the classifier space. For instance, the conjunctive composition function $g^{\land}\func$, given two concepts as input like ``Socks'' and ``Red'' represented in the classifier space by $w_s, w_r \in \reals^D$, should compute the classifier $w_{s~\land~r} \in \reals^D$ that recognizes when both concepts are present in a image simultaneously. Similarly, the functions $g^{\lor}\func$ and $g^{\neg}\func$ should compute the disjunction and negation in classifier space, respectively.

We also observe that some of these composition functions can be defined analytically or in terms of other composition functions. More specifically, the negation consists of just inverting the separating hyperplane and the disjunction can be derived using De Morgan's laws. Then, we propose to implement these functions as
\begin{equation}
\begin{aligned}
g_\theta^{\land}(w_a, w_b) &= \text{Neural Network}(w_a, w_b) \\
g^{\neg}(w) &= -w \\
g^{\lor}\left(w_a, w_b \right) &= g^{\neg}\left(g^{\land}\left(g^{\neg}\left(w_a\right), g^{\neg}\left(w_b\right)\right)\right),
\end{aligned}
\end{equation}
where the conjunctive composition $g_\theta^{\land}\func$ is a neural network learned from data and $\theta$ are the learnable parameters.\footnote{Equivalently, we could have defined $g^{\lor}$ by the neural network and $g^{\land}$ using De Morgan's laws.} Therefore, the learning of function $f_\theta\func$ is decomposed on the learning of these composition functions. 

Following these ideas, let us define a subset of training expressions $\{e_k\}_{j=1}^K \subset \E$ composed by composition rules $\ops$ and primitives $\P$. Note that such a subset is much smaller than all possible expressions that can be formed by composing these primitives. Likewise, we define a set of training images $\lbrace (x_i, y_i) \mid x_i \in \I, y_i \in \{0, 1\}^K \rbrace_{i=1}^N$ with the ground-truth label $y_{ij}$ denoting whether the image $x_i$ is a positive example for the expression $e_j$. Then, learning the function $f_\theta\func$ can be defined as,
\begin{multline}
\label{eq:obj}
\minimize{\theta, \phi} ~ \frac{1}{KN} \sum_{j=1}^{K} \sum_{i=1}^{N} ~ \alpha_1 \Delta\left(f_\theta(e_j)^T h_\phi\left(x_i\right), y_{ij} \right) \\ + \frac{\alpha_2}{2} \norm{f_\theta(e_j)}_2^2 + \alpha_3 R(\theta),
\end{multline}
where $\Delta(\cdot,\cdot)$ is a classification loss function, $R\func$ is some regularization function and $\{\theta,\phi\}$ is the set of learnable parameters. We also have the hyper-parameters $\alpha \in \reals^3$ which controls how our model correctly fit the training data ($\alpha_1$), regularizes for training expressions ($\alpha_2$), and for unknown expressions ($\alpha_3$). The idea is to learn how to synthesize classifiers that correctly classify images according to the input expressions even if the expressions had not been seen during training.

It is important to note that such a formulation aims to explore semantic similarity on classifiers space and the visual compositionality principle in order to make our learning problem easier to solve. We use a relative small subset of expressions to learn our proposed mapping function and rely on the classifier similarity to generalize for unknown expressions. Likewise, we explore visual compositionality by decomposing training expressions in simpler expressions and jointly learning the composition functions.

\subsection{Inference}
As alluded to above, our main goal is to produce classifiers for boolean expressions of primitives. These expressions can be represented by a tree where composition rules are nodes and primitives are leaves. Thus, our inference consists of parsing the expression tree in post-order and applying the composition functions accordingly in order to end up with the final classifier just after parsing the root. 

Then, we can compute the classifier score for an image given an expression by:
\begin{equation}
s = f_\theta(e)^T h_\phi(x)
\end{equation}
This score reflects the compatibility between the expression and the image. We want this score to be high only if the image contains the complex concept described by the expression $e$ and low otherwise. As an example, for the expression ``$S \land (B \lor R) \land \neg H$'' we want the score $s$ to be high only for images containing blue or red socks without holes and want it to be low for images containing any other concept.

\subsection{Model and Implementation Details}
We propose to implement the conjunctive composition function $g_\theta^{\land}\func$ and the feature extractor $h_\phi\func$ as a multilayer perceptron (MLP) \citep{Haykin:2009} network and VGG-16 convolutional neural network \citep{Simonyan:2014} respectively. We represent images with 4096-dimensional feature vectors computed by the FC6-layer of VGG-16 network pretrained on ImageNet~\citep{ILSVRC14}. Consequently, the primitives are represented by 4097-dimensional vector obtained from training linear SVMs on these features. Since the bias can be implemented by adding a $+1$ fixed feature to image representation vectors, $g_\theta^{\land}\func$ is a MLP network that have $(2 \times 4097)$ inputs and two fully connected layers with outputs of size $(1.5 \times 4097)$ and $(4097)$, respectively. We use the LeakyReLU non-linearity, with slope set to $0.1$, in between the layers and linear activation on the outputs. \figref{fig:model} shows our neural network architecture in details. 

During training, we approximate the objective \eqnref{eq:obj} by batches of 32 expressions, 5 positive and 5 negative images for each expression sampled uniformly. We first train our neural algebra of classifiers module alone during 50 epochs, then we finetune the features jointly during 10 epochs more. Since the primitives are represented by linear SVM classifiers, we decide to use the hinge loss,
\[
\Delta\left(s_{ij}, y_{ij}\right) = \max(1 - y_{ij} s_{ij})
\]
where $s_{ij}$ is the score assigned to the image $x_i$ by the classifier predicted for the expression $e_j$. In addition, we use the standard $\ell_2$ regularization in the network weights as our regularization function $R\func$.

\section{Experiments}
We now evaluate the performance of our method and compare against several baselines. We first describe the experimental setup, datasets, metrics, and baselines used in our experiments. Then, we analyze how effectively our model can compose classifiers for simple and arbitrary compositions of concepts in addition to presenting a qualitative evaluation of our method.

\subsection{Experimental Setup \label{sec:exp_setup}} 
\begin{table*}[t!h]
\centering
\caption{Evaluating known/unknown disjunctive and conjunctive expressions on the CUB-200 Birds dataset.}
\label{tab:single_op:cub200}
\resizebox{1\textwidth}{!}{\begin{tabular}{l|ccc|ccc|ccc|ccc}
  	   	     & \multicolumn{6}{c|}{Disjunctive Expressions} & \multicolumn{6}{c}{Conjunctive Expressions} \\
       		 &  \multicolumn{3}{c|}{Known Exp.}   &    \multicolumn{3}{c|}{Unknown Exp.}  &  \multicolumn{3}{c|}{Known Exp.}   &    \multicolumn{3}{c}{Unknown Exp.}  \\
Metrics  & MAP & AUC & EER & MAP &  AUC & EER & MAP & AUC & EER & MAP &  AUC & EER \\             
             
\hline
Chance & 39.70 & 50.00 & 50.0 & 40.60 & 50.00 & 50.0 & 4.55 & 50.0 & 50.0 & 4.59 & 50.0 & 50.0 \\   

Supervised & 65.25 & 74.76 & 31.58 & - & - & - & 22.87 & 78.02 & 29.69 & - & - & - \\   

Independent & 58.73 & 68.39 & 36.76 & 60.66 & 69.28 & 36.10 & 17.23 & 77.22 & 29.94 & 19.16 & 78.00 & 29.28 \\   
\hline
\textbf{Neural Alg. Classifiers}  &\textbf{ 70.10} & \textbf{77.36} & \textbf{29.44} & \textbf{71.18} & \textbf{77.76} & \textbf{29.04} & \textbf{23.09} & \textbf{81.54} & \textbf{26.36} & \textbf{23.87} & \textbf{81.98} & \textbf{25.85} \\   
\hline
\end{tabular}}
\end{table*}

\begin{table*}[t!h]
\centering
\caption{Evaluating known/unknown disjunctive and conjunctive expressions on the AwA2 dataset.}
\label{tab:single_op:awa2}
\resizebox{1\textwidth}{!}{\begin{tabular}{l|ccc|ccc|ccc|ccc}
  	   	     & \multicolumn{6}{c|}{Disjunctive Expressions} & \multicolumn{6}{c}{Conjunctive Expressions} \\
       		 &  \multicolumn{3}{c|}{Known Exp.}   &    \multicolumn{3}{c|}{Unknown Exp.}  &  \multicolumn{3}{c|}{Known Exp.}   &    \multicolumn{3}{c}{Unknown Exp.}  \\
Metrics  & MAP & AUC & EER & MAP &  AUC & EER & MAP & AUC & EER & MAP &  AUC & EER \\             
             
\hline
Chance & 53.19 & 50.0 & 50.0 & 53.04 & 50.0 & 50.0 & 18.77 & 50.0 & 50.0 & 21.17 & 50.0 & 50.0 \\   

Supervised & 97.47 & 97.20 & 8.13 & - & - & - & 94.90 & 98.53 & 6.00 & - & -  & - \\   

Independent & 97.28 & 97.12 & 8.70 & 97.86 & 97.58 & 6.77 & 93.95 & 98.13 & 6.80 & 93.90 & 97.87 & 7.36 \\   
\hline
\textbf{Neural Alg. Classifiers}  & \textbf{98.84} & \textbf{98.67} & \textbf{5.84} & \textbf{99.05} & \textbf{98.91} &\textbf{ 5.24} & \textbf{95.95} & \textbf{98.79} & \textbf{5.29} & \textbf{96.50} & \textbf{98.81} & \textbf{5.34} \\   
\hline
\end{tabular}}
\vspace{-5px}
\end{table*}

We are interested in the task of predicting whether a given image contains the complex concept described by a boolean expression of primitives which may not have any training data. Towards this end, we first define two disjoint sets of boolean expressions of primitives named ``training expressions'' and ``test expressions'' and three disjoint sets of images named ``training images'', ``validation images'' and ``test images''. Second, we learn the primitive representation, train our model and baselines using training images and training expressions. Then, we evaluate the performance of our method and baselines classifying images on the validation set according to training expressions, named ``known expressions performance'', and classifying images on the test set according to test expressions, named ``unknown expressions performance''. The former suggests how well a model learns to compose classifiers and the latter how well a model generalizes for expressions not seen in training.

\vspace{-2px}
\paragraph{Datasets.} We use the CUB-200 Birds (CUB-200) \citep{CUB_200_2011} and Animal With Attributes 2 (AwA2) \citep{Xian:CVPR17} datasets in our experiments. Since none of these datasets were designed for our purpose, we split these datasets in order to perform controlled experiments. First, we compute all possible binary conjunctive and disjunctive expressions of primitives and filter out the ones that do not have reasonable amount of positive and negative images. Then, we randomly split the images between train, validation, and test images making sure that every expression and primitive have reasonable amounts of positive and negative samples in each image split. As a result, we create approximately $3k$ training expressions and $1k$ test expressions using $250$ primitives for the CUB-200 dataset, while we create approximately $1.5k$ training and $600$ test expressions using $80$ primitives for the AwA2 dataset. In order to make easier to reproduce our results, the experiment code and these data splits are available in the first author's homepage.

\vspace{-2px}
\paragraph{Metrics.} A boolean expression of primitives defines a binary classification problem where images are classified as relevant or irrelevant for the visual concept described. Therefore, we use well-known evaluation metrics of image retrieval and binary classification. More specifically, we use the mean average precision (MAP), area under the ROC curve (AUC) and equal error rate (EER). We compute these metrics globally in order to take the data imbalance in account since some expressions are naturally rarer than others.

\vspace{-2px}
\paragraph{Baselines.} We compare our method to several baselines in order to evaluate empirically how well we can compose classifiers for complex concepts: 
\begin{itemize}
\item Chance: This is an empirical lower bound for the problem and consists of assigning random scores for image and expression pairs.
\item Supervised: This is an empirical upper bound for the problem and consists of training SVMs for every training expression. Thus, it is a fully supervised approach which can not be extended for unknown expressions. Therefore, we just report its performance for known expressions.
\item Independent Classifiers: This baseline assumes that visual concepts are independent events and uses basic probability rules to estimate the probability of a complex concept being depicted in an image. They are defined according to the following rules,
\begin{equation}
\begin{aligned}
p(v_1~\land~v_2) &= p(v_1)p(v_2) \\
p(v_1~\lor~v_2) &= p(v_1) + p(v_2) - p(v_1)p(v_2)\\ 
p(\neg~v) &= 1 - p(v) \\
\end{aligned}
\end{equation}
where $p(v)$ is the probability of a given image has the primitive $v$ estimated by the classifier $w_v$. Note that in order to estimate these probabilities we calibrate the learned SVMs using a small held-out subset of the training images $(\approx 10\%)$ and Platt's calibration method \citep{Platt:1999}.
\end{itemize}

\subsection{Simple Binary Expressions}

In this experiment, we focus on evaluate how well our model can learn to compose classifiers for simple binary conjunctive and disjunctive expressions. We follow the procedure explained in \secref{sec:exp_setup} and evaluate our model and baselines on both cases separately. We do not report the result with simple negative expressions since it is a trivial mapping in classifier space as explained in \secref{sec:method}.

We present the results for our methods and baselines on the CUB-200 and AwA2 datasets in \tabref{tab:single_op:cub200} and \tabref{tab:single_op:awa2} respectively. As expected, the supervised method presents good performance on both types of expressions but it is limited to expressions known at training phase. Thus, it can not be used in large scale recognition problems where the number of complex concepts that can be composed is very large. 

On the other hand, the independent approach seems to be a strong baseline. It produces slightly worse results than the supervised approach for known expression, mainly on conjunctive expressions, while can classify images according to unknown expressions. However, we note that such a performance is due to the high accuracy of the primitive classifiers, it can reach the AUC of $\approx 85\%$ for the CUB-200 and $\approx 95\%$ for the AwA2 when classifying validation and test images according to primitive concepts. Then, its performance should decrease drastically in more challenging datasets.

However, our method shows significant superior performance on every setting on both datasets. For instance, the proposed method reaches improvements around $10\%$ for disjunctive expressions and $5\%$ for conjunctive expressions in the CUB-200 dataset. In fact, it is able to surpass the supervised methods on known expression since it allows to learn specific features for complex compositions in addition to reason about correlations between primitives. It is also important to mention that our hypothesis of implementing the disjunctive composition function as the combination of the negation and conjunction according to the De Morgan's laws is verified, since we reach similar performance, when we train a specific MLP network for disjunctive expressions.

Despite the differences highlighted in \secref{sec:relw}, we acknowledge the similarity between the transformation function proposed by \citet{Misra:Composing:CVPR2017} and our AND composition function. More specifically, we both learn an MLP, but we use different network architectures and optimize different objectives. Then, we evaluate their model in our simple binary conjunctive expression experiment. Despite their model having approx. 2.7x more learnable parameters, it performs slightly worse than our AND composition (around 1\% in all metrics used) which demonstrates the efficiency of our architecture and loss function.

\subsection{Complex Expressions}
\begin{figure*}[t!h]
	\begin{center}		
		\includegraphics[width=0.33\textwidth]{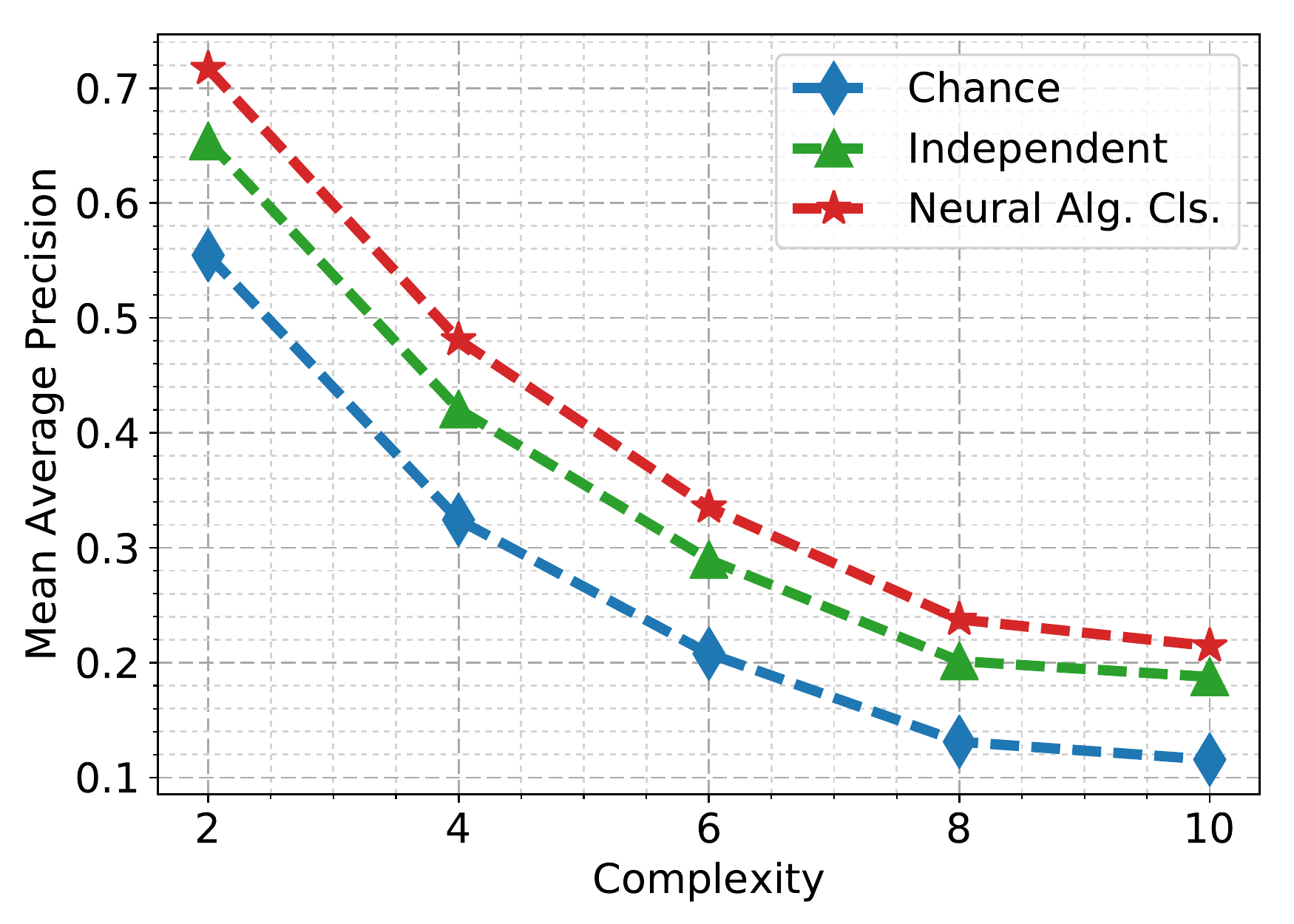}
		\includegraphics[width=0.33\textwidth]{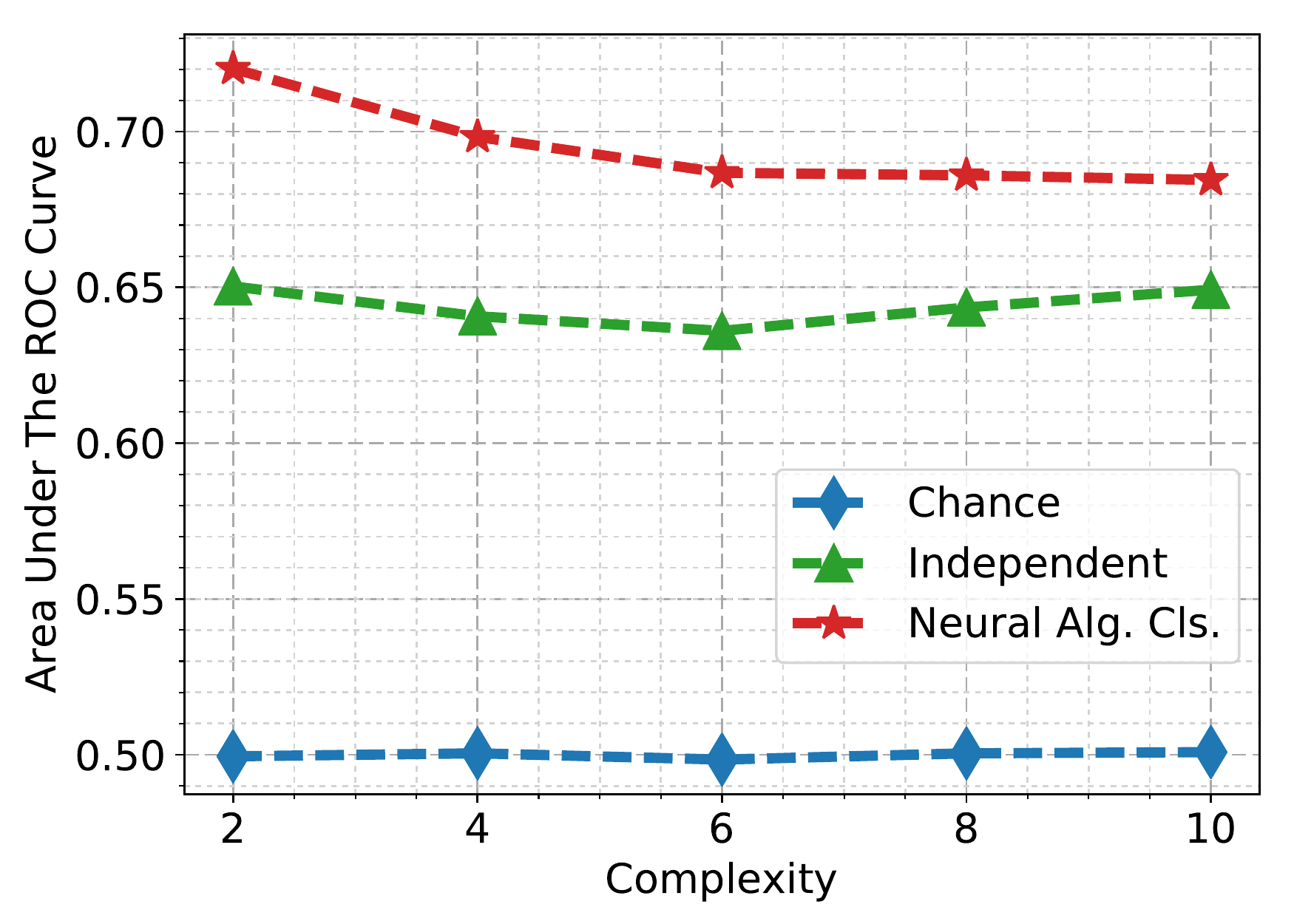}
		\includegraphics[width=0.33\textwidth]{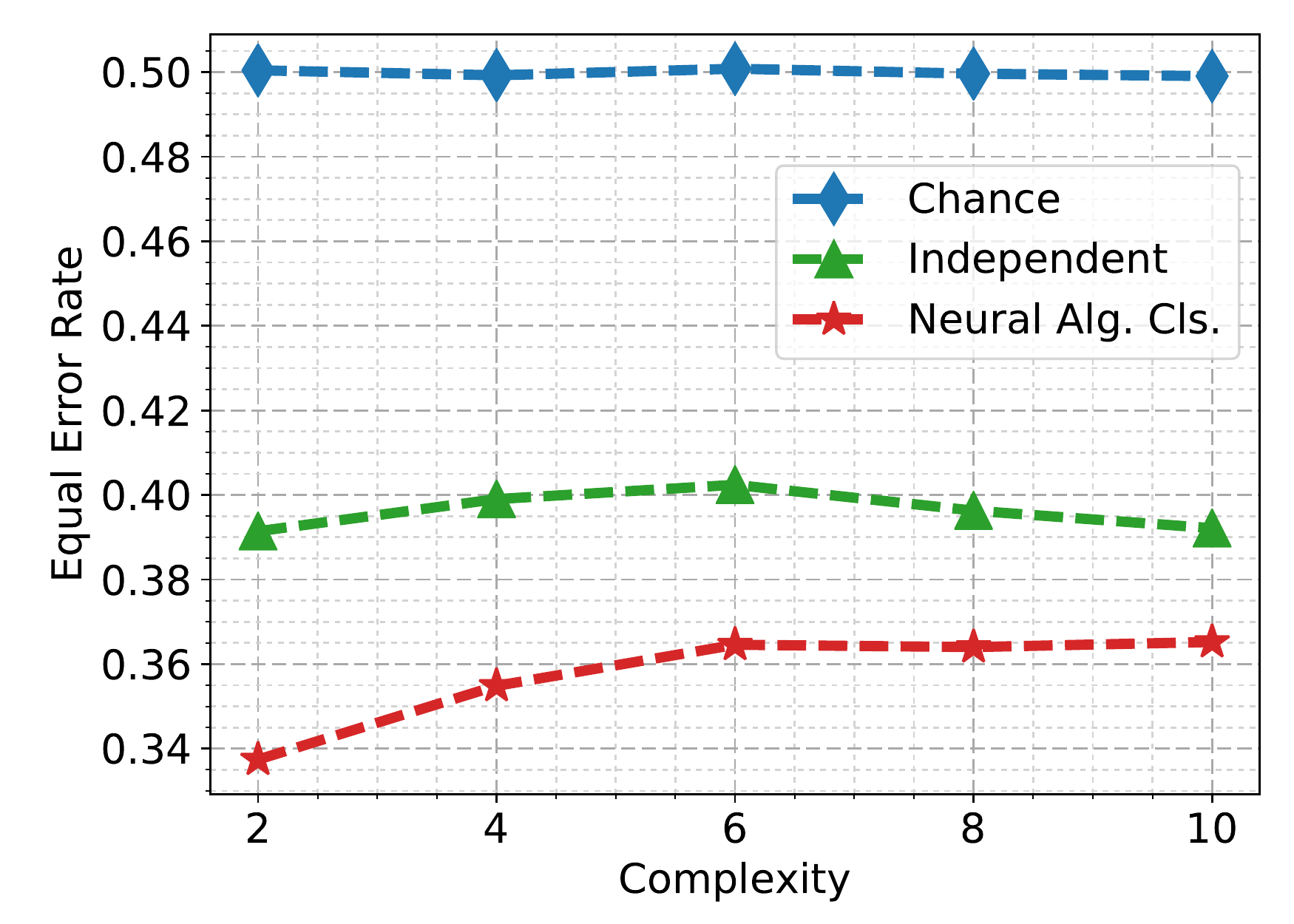}\\
		\includegraphics[width=0.33\textwidth]{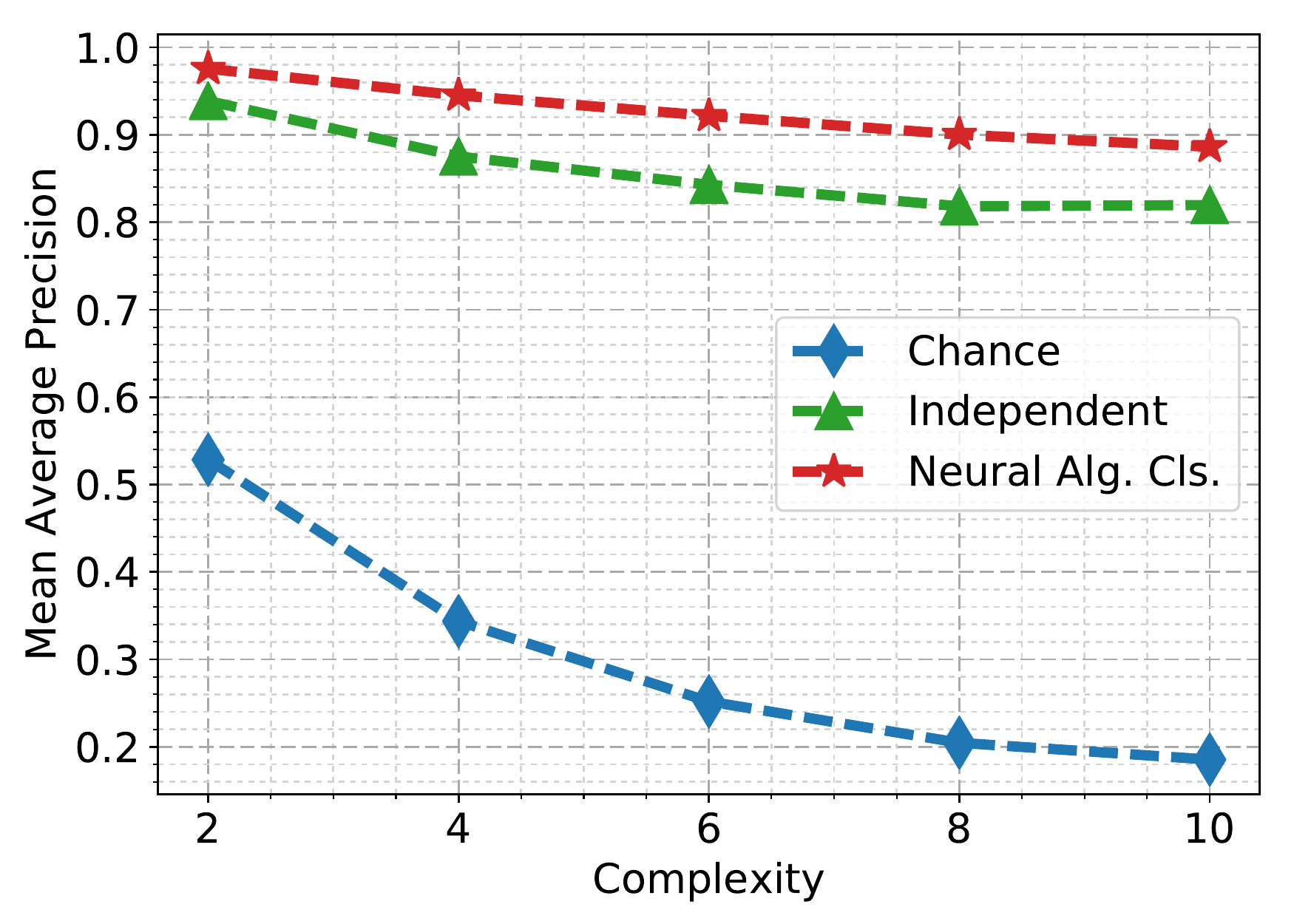}
        \includegraphics[width=0.33\textwidth]{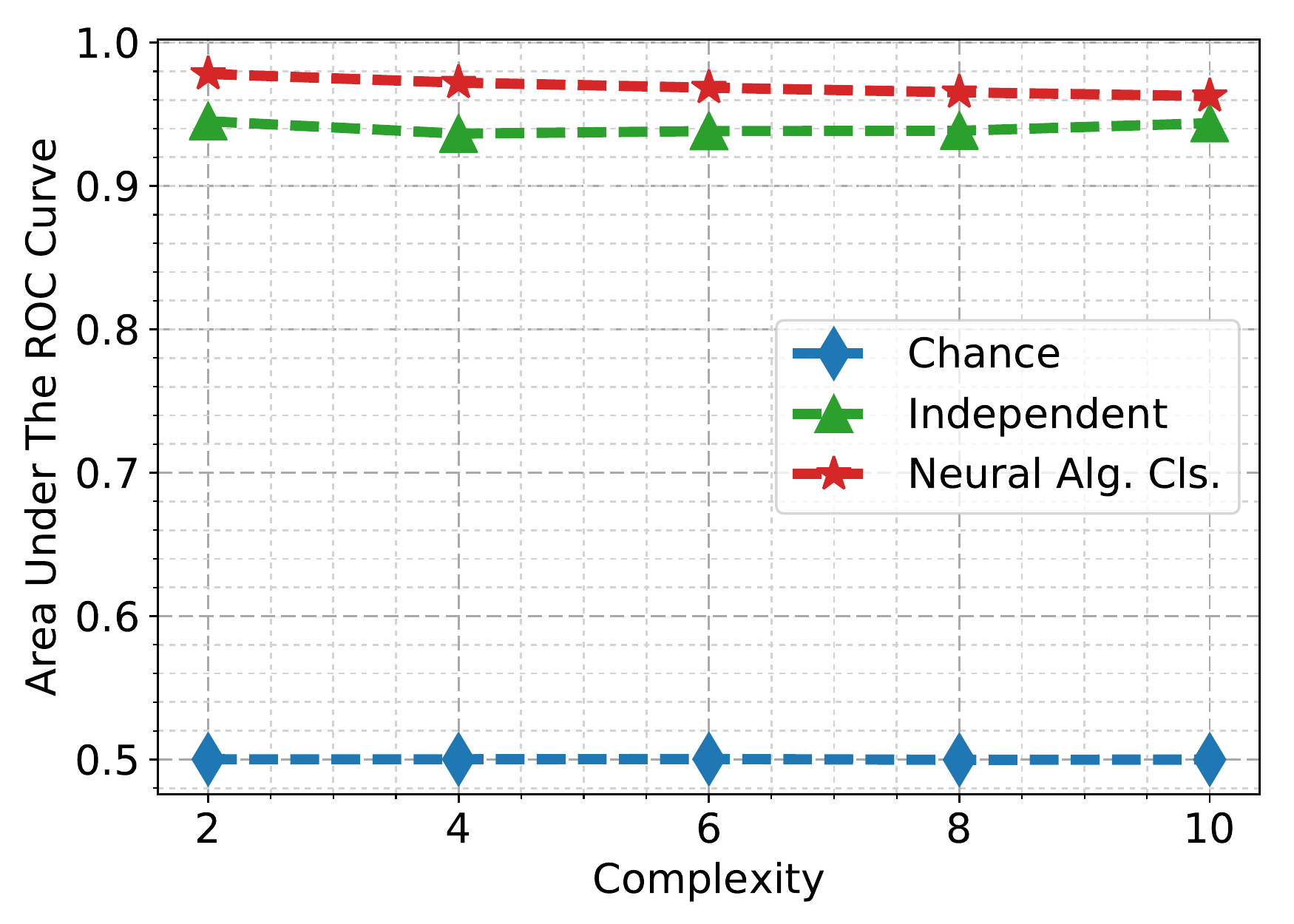}
        \includegraphics[width=0.33\textwidth]{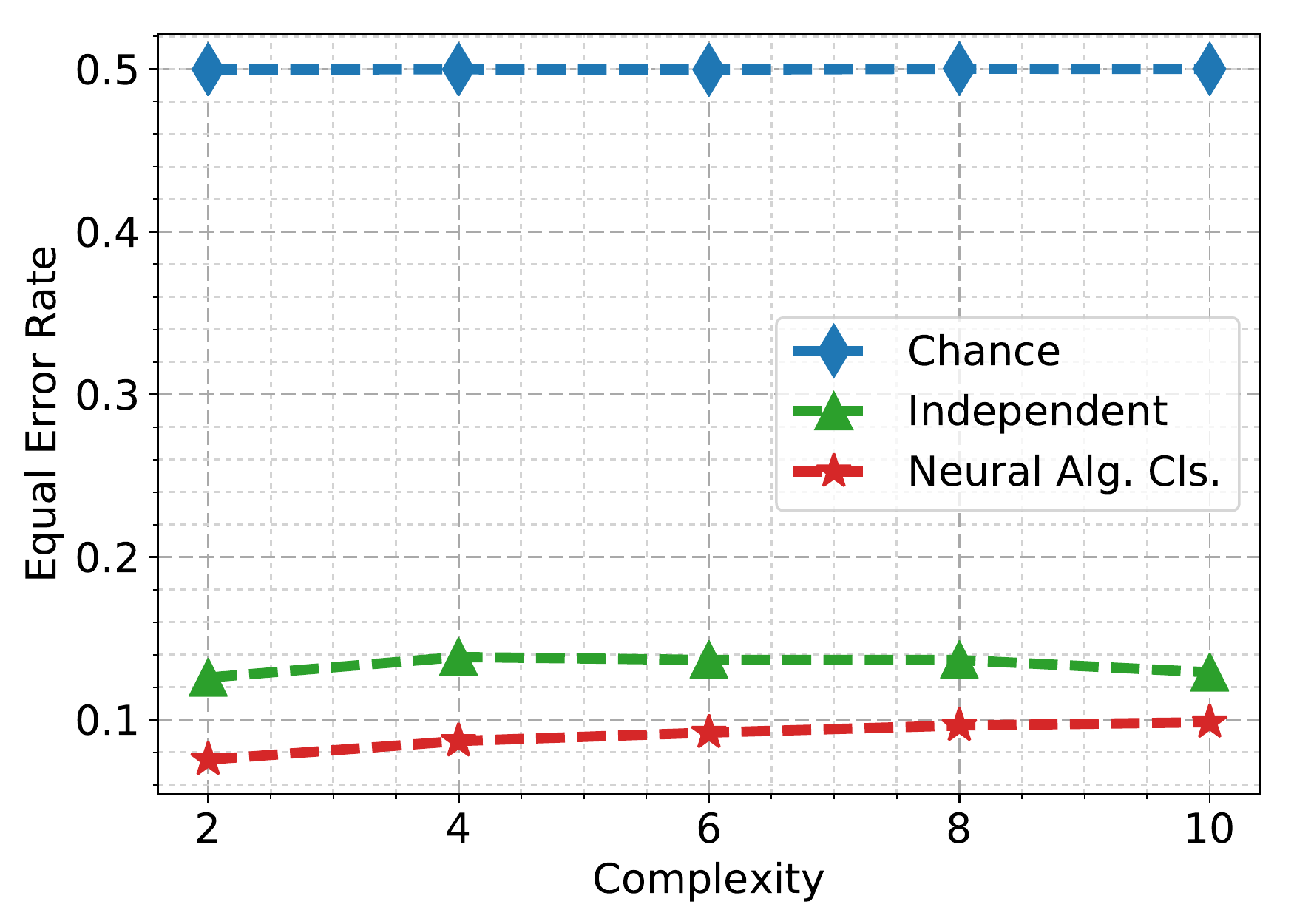}        
	\end{center}
	\vspace{-5px}
\caption{Performance of the proposed method and baselines on classifying images according to unknown expressions of different complexity described in conjunctive normal form (CNF). The first row presents the results for CUB-200 dataset, while the second row presents the results for AwA2 dataset. In the first column the performance in measured in terms of mean average precision (higher is better), while the second column reports the area under the ROC curve (higher is better), and the third column reports the equal error rate (lower is better). }
\label{fig:multiple_op}
\vspace{-5px}
\end{figure*}

From previous experiments, we can conclude that our model is able to learn composition rules for simple binary expressions. However, we still need to show that these models are suitable for arbitrary expressions. According to boolean algebra, every boolean expression can be written in generic forms such as Normal Disjunctive From (NDF) and Normal Conjunctive form (NCF). The former consists of an OR of ANDs, e.g., $(p_1 \land q_1) \lor (p_2 \land q_2) \lor \ldots \lor (p_c \land q_c)$, and the latter consists of an AND of ORs, e.g., $(p_1 \lor q_1) \land (p_2 \lor q_2) \land \ldots \land  (p_c \lor q_c)$ where $p$ and $q$ are visual primitives which may appear negated and $c$ is the number of simple terms in those expressions. From the visual recognition perspective, $c$ can be seen as an indicator of the complexity of an expression since long expressions usually defines more specific visual concepts than short expressions. For instance, $(Blue \lor Red) \land Socks \land (\neg Holes)$ is a more specific visual concept than any of its subexpressions such as $((Blue \lor Red) \land Socks)$ and $(Socks \land (\neg Holes))$.

Since it is straight forward to convert any expression for both normal forms \citep{Monk:BoolAlg:1989}, we decide to examine the performance of our method and baselines on complex expressions in the normal conjunctive form. Towards this end, we randomly generate 1k test CNF expressions of complexity 2, 4, 6, 8, 10 from simple unknown disjunctive expressions. In order to avoid normalization issues when combining linear classifiers produced by our method and the primitives classifiers, we finetune our method using training images and CNF expressions of complexity 4 formed from known simple disjunctive expressions. Then, we use our method and baselines to classify test images according to the sampled CNF expressions of different complexities. Again, the finetune and test expression sets are disjoint as well as the training and test image sets. We also do not evaluate the supervised baseline because we do not have training images for the test expressions.

In \figref{fig:multiple_op}, we plot baselines and our method performance in terms of mean average precision, area under the ROC curve and equal error rate on CNF expressions of different complexities composed by unknown simple binary expressions. As expected, the performance of all evaluated methods decrease as we increase the complexity of the test expressions. This is more noticeable in our method which stabilizes for complexity greater or equal to 6. However, we consistently outperform the baselines on classifying images according to expressions of different complexities in both datasets. 

\subsection{Qualitative Evaluation}
\begin{figure*}[t!h]
	\begin{center}		
		\includegraphics[width=\textwidth]{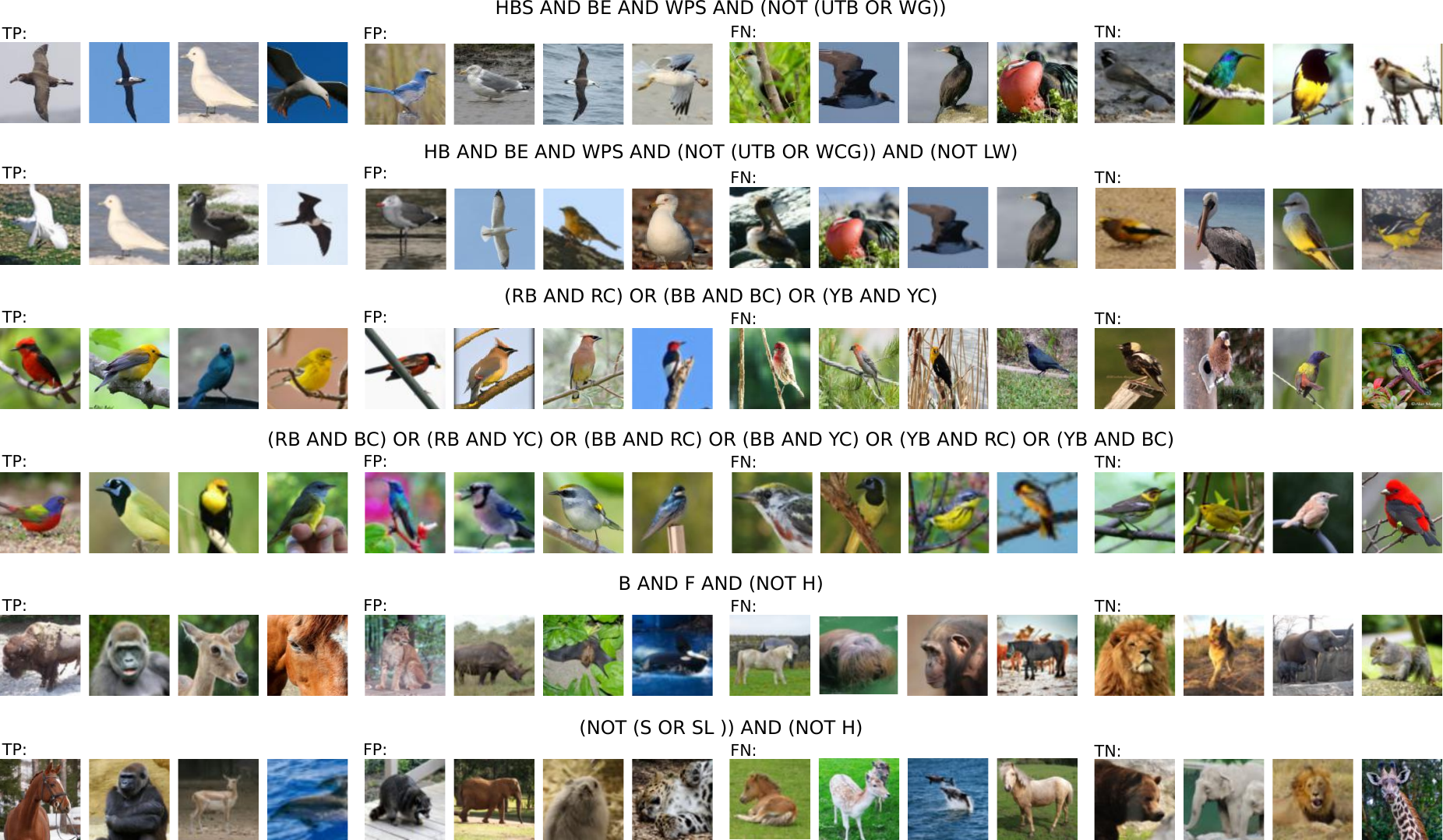}
	\end{center}
    \vspace{-5px}
	\caption{Randomly selected true positives (TP), false positives (FP), false negatives (FN) and true negatives (TN) test images classified according to manually selected unknown expressions of the following visual primitives: hooked beak (HB), black eyes (BE), solid wings pattern (WPS), long wings (LW), blue breast (BB), red breast (RB), yellow breast (YB), blue crown (BC), red crown (RC), yellow crown (YC), big (B), fast (F), hunter (H), small (S) and slow (SL).}
	\label{fig:demo}
    \vspace{-5px}
\end{figure*}

We now evaluate the proposed method qualitatively by visualizing the classification results of some interesting expressions. More specifically, we classify the test images by scoring them according to manually picked unknown expressions and thresholding using the equal error rate threshold. In \figref{fig:demo}, we show some randomly selected true positives (TP), false positives (FP), false negatives (FN) and true negatives (TN) for every selected expression. 

Looking back to our motivational example and analyzing the ground truth of CUB-200 dataset, we can state that albatrosses and gulls are birds with hooked beak (HB), black eyes (BE), solid wings pattern (WPS) which do not have black upper tail (UTB) or gray wings (WG). We examine such a statement in the first row of \figref{fig:demo} by analyzing the classification results produced by our method for the respective boolean expression of these primitives. We note that most of the positive predictions are from different species of albatrosses and gulls. Furthermore, long wings (LW) is a good visual feature to discriminate albatrosses from gulls. Then, we add such a term in the boolean expression and note the predominance of gulls in the predicted positive examples in the second row of \figref{fig:demo}. This example shows qualitatively that our approach is able to group and discriminate objects according to different visual features.

In addition, we can also use our method to find specific combinations of visual features. For instance, consider the following visual features: blue breast (BB), red breast (RB), yellow breast (YB), blue crown (BC), red crown (RC) and yellow crown (YC). In the third row of \figref{fig:demo}, we are looking for birds that have the breast and crown of the same color which could be blue, red or yellow. While in the fourth row of \figref{fig:demo}, we aim for a more specific combinations of these visual primitives like birds that have different breast and crown color. We can note that the predicted positives are predominately unicolor in the former expression, while they are more colorful in the latter one. Furthermore, the false positives usually present part of the desired composition of visual primitives which is perhaps a consequence of the compositional principle.

From the perspective of boolean algebra, two equivalent expressions must have the same truth table. Translating to our context, we can say that two equivalent composition of primitives should have similar classification results. In order to demonstrate such a property, we express the set of big (B) and fast (F) animals that are not hunter (H) in two different ways using De Morgan's Laws: (B AND F) AND (NOT H) and (NOT (S OR SL)) AND (NOT H) where small (S) and slow (SL) are the opposite concepts of fast and big respectively. As we can see in the last two rows of \figref{fig:demo}, the positive and negative predictions have basically instances from the same classes such as gorillas, deers, horses and dolphins for the positives while elephants, tigers and lions for the negatives. Therefore, our proposed method spans an algebra of visual primitives where complex visual concepts can be described by different compositions.

\section{Conclusion}
In this paper, we tackled the problem of learning to synthesize classifiers for complex visual concepts expressed in terms of visual primitives. We formulated such a problem as an algebra of classifiers where the composition rules are learned from data and complex visual concepts are expressed by boolean expressions of primitives. Through a variety of experiments, we show that our framework can synthesize accurate classifiers for known expressions, and generalize to arbitrary unknown expressions. It consistently outperforms the baselines across different metrics and datasets. Besides, we demonstrate qualitatively different queries that can be answered by our model. 


\small{
\smallskip \noindent \textbf{Acknowledgements:} This research was supported by the Australian Research Council (ARC) through the Centre of Excellence for Robotic Vision (CE140100016) and was undertaken with the resources from the National Computational Infrastructure (NCI), at the Australian National University (ANU).}

{\bibliography{short,nbac_bib}}

\end{document}